\documentclass[letterpaper]{article} 
\usepackage{aaai2026}  
\usepackage{tabularx}
\usepackage{times}
\usepackage{helvet}  
\usepackage{courier}  
\usepackage[hyphens]{url}  
\usepackage{graphicx} 
\usepackage{natbib}  
\usepackage{caption} 
\usepackage{algorithm}
\usepackage{algorithmic}
\usepackage{booktabs}

\usepackage{newfloat}
\usepackage{listings}
\DeclareCaptionStyle{ruled}{labelfont=normalfont,labelsep=colon,strut=off} 
\floatstyle{ruled}
\newfloat{listing}{tb}{lst}{}
\floatname{listing}{Listing}
\title{AdiBhashaa: A Community-Curated Benchmark for Machine Translation into Indian Tribal Languages}

\author{
    Pooja Singh\textsuperscript{\rm 1},
    Sandeep Kumar\textsuperscript{\rm 1,2}
}

\affiliations{
    \textsuperscript{\rm 1}Department of Electrical Engineering, Indian Institute of Technology Delhi \\
    \textsuperscript{\rm 2}Yardi School of Artificial Intelligence, Indian Institute of Technology Delhi \\
    New Delhi, India \\
    \texttt{eez228470@iitd.ac.in, ksandeep@iitd.ac.in}
}

\begin{document}

\maketitle

\begin{abstract}

Large language models and multilingual machine translation (MT) systems increasingly drive access to information, yet many languages of the tribal communities remain effectively invisible in these technologies. This invisibility exacerbates existing structural inequities in education, governance, and digital participation. We present AdiBhashaa, a community-driven initiative that constructs the first open parallel corpora and baseline MT systems for four major Indian tribal languages-Bhili, Mundari, Gondi, and Santali. This work combines participatory data creation with native speakers, human-in-the-loop validation, and systematic evaluation of both encoder-decoder MT models and large language models. In addition to reporting technical findings, we articulate how AdiBhashaa illustrates a possible model for more equitable AI research: it centers local expertise, builds capacity among early-career researchers from marginalized communities, and foregrounds human validation in the development of language technologies.

\end{abstract}

\section{Introduction}

Despite substantial advances in multilingual NLP, coverage of languages spoken in the tribal communities remains highly uneven. India has more than 460 tribal languages and 71 distinct tribal mother tongues, spoken by over 100 million people, yet the vast majority of these languages have no digitized resources, no MT systems, and minimal representation in foundation models~\cite{xaxa2005politics}. This digital exclusion mirrors and amplifies long-standing social and economic marginalization.
We address this gap by focusing on four widely spoken tribal languages: \emph{Bhili} (Indo-Aryan), \emph{Mundari} and \emph{Santali} (Austroasiatic, Munda branch), and \emph{Gondi} (Dravidian, Central Dravidian subgroup) as shown in Table~\ref{tab:tribal-stats}~\cite{census2011_st15_bihar}. Together, these languages represent diverse linguistic families, scripts, and geographic regions across central and eastern India. Yet none appear as distinct units in common MT benchmarks or pretraining corpora.
Our goals are twofold. First, we seek to create high-quality, openly available resources that enable MT between these tribal languages and dominant languages such as Hindi and English. Second, we aim to develop a methodological template for collaborative, community-led AI research that directly involves speakers of underrepresented languages as co-designers, not merely data annotators. We argue that such practices are essential to ``under-represented languages'' in a substantive and sustainable manner.

\section{Community-Driven Corpus Creation}
\subsection{Languages, Scripts, and Domains}

We construct a 80,000 sentence parallel corpus comprising Hindi–tribal data for four low-resource languages, Bhili, Mundari, Gondi, and Santali, allocating 20,000 sentence pairs per language. Source sentences are drawn from domains that are both linguistically diverse and socially consequential for underrepresented language communities, including school and higher-education materials, government and civic information, political speeches, news and mass media, healthcare messaging, and everyday social communication.
Bhili, Mundari, and Gondi are primarily written in Devanagari, whereas Santali uses the Ol Chiki script. This combination of shared and low-resource scripts enables us to study two complementary challenges: (i) cross-script transfer from a high-resource language (Hindi) to closely related writing systems, and (ii) modeling for underserved scripts such as Ol Chiki, which have limited digital infrastructure and minimal representation in existing large-scale corpora.


\begin{table}[t]
\small
\centering
\setlength{\tabcolsep}{3pt} 
\begin{tabularx}{\columnwidth}{l l l X}
\toprule
\textbf{Language} & \textbf{Family} & \textbf{Speakers} & \textbf{Spoken in States} \\
\midrule
Santali  & Austroasiatic & 7.6 M &
Jharkhand, West Bengal, Odisha, Assam \\
Bhili    & Indo-Aryan            & 10.4 M &
Madhya Pradesh, Rajasthan, Gujarat \\
Mundari   & Austroasiatic  & 1.1 M &
Jharkhand, Odisha, West Bengal, Assam \\
Gondi  & Dravidian   & 3.0 M &
Madhya Pradesh, Maharashtra, Chhattisgarh, Telangana,\\
               &                       &      &
Andhra Pradesh, Odisha \\
\bottomrule
\end{tabularx}
\caption{Statistics of selected tribal languages (2011 Census of India; speaker counts approximate).}
\label{tab:tribal-stats}
\end{table}

\subsection{Participatory Translation Workflow}

To ensure linguistic quality, cultural fidelity, and community ownership, we adopt a participatory, human-in-the-loop workflow rather than relying solely on crowdsourcing or model-generated text. For each language, we collaborate with native-speaking translators who are experienced teachers or language activists with deep community engagement.

Our workflow proceeds in three structured stages:

\begin{itemize}

\item \textbf{Source curation.}
Researchers compile Hindi sentences from public, educational, and civic sources, prioritizing cultural appropriateness, clarity, and likely downstream utility (e.g., education, public services).

\item \textbf{Community translation.}
Native-speaking translators render the sentences into their languages with an emphasis on semantic adequacy and naturalness. Translators are encouraged to adapt idiomatic expressions, examples, or cultural references when doing so improves local comprehensibility and preserves community norms.

\item  \textbf{Independent validation.}
A separate group of validators, often students or early-career researchers from the same communities, reviews each sentence pair for adequacy, fluency, and cultural sensitivity. Disagreements are resolved through structured discussion sessions that double as informal training in annotation practice, linguistic analysis, and MT error identification.
\end{itemize}

This workflow produces a corpus that is not only technically robust for machine translation but also socially grounded in community linguistic practices. 


\section{Modeling and Evaluation}
\subsection{Baselines and Experimental Setup}

We establish baseline systems for Hindi/English $\leftrightarrow$ tribal translation using both neural machine translation (MT) models and LLMs. On the MT side, we evaluate multilingual encoder-decoder architectures such as NLLB-200~\cite{costa2022no}, mT5~\cite{xue2021mt5}, and IndicTrans2~\cite{galaindictrans2}, which are pre-trained on many languages but do not explicitly include Bhili, Mundari, Gondi, or Santali as distinct identifiers. On the LLM side, we assess representative open and closed-source models in zero-shot and few-shot in-context settings such as Bloom~\cite{muennighoff2023crosslingual}, Gemini 2.5 Flash~\cite{comanici2025gemini}, Gpt-4o-mini~\cite{hurst2024gpt} and many others. For each translation direction, MT models are fine-tuned on 95\% of the available parallel data, with 5\% held out for evaluation. English counterparts of Hindi sentences are produced with a IndicTrans2 model and filtered via manual inspection of sampled outputs. We report chrF++ and sentence-level BLEU, and complement these with human evaluation by native speakers on a curated subset of test sentences.



\subsection{Quantitative Results}

Three clear empirical patterns emerge:

\begin{enumerate}
    \item \textbf{Fine-tuned multilingual MT outperforms zero-shot.} Fine-tuning multilingual encoder--decoder models on the AdiBhashaa corpus yields consistent, substantial gains over zero-shot performance for all four languages, showing that a carefully curated, community-produced dataset of modest size provides sufficient signal for models to learn previously unseen languages.

\item \textbf{Asymmetry between tribal $\rightarrow$ high-resource and high-resource $\rightarrow$ tribal.} Translation into high-resource languages (tribal $\rightarrow$ Hindi/English) systematically outperforms translation into the tribal languages according to both automatic metrics and human judgments. This reflects known difficulties in generating morphologically rich, low-frequency vocabulary on the target side and underscores the need to expand both the volume and domain coverage of tribal-language data.

    \item \textbf{Few-shot LLMs are competitive but lag behind fine-tuned MT.} LLMs with few-shot prompting provide a strong baseline when fine-tuning is infeasible, but generally underperform dedicated fine-tuned MT models, especially for directions involving Santali in Ol Chiki. Task-specific adaptation thus remains essential for tribal languages that are largely absent from pretraining corpora.
\end{enumerate}

\section{Broader Implications}

AdiBhashaa offers broader lessons for building inclusive AI ecosystems.


\paragraph{Capacity building through research practice.}
AdiBhashaa serves as an entry point into AI research for students and early-career scholars from tribal and other low-resource backgrounds. Regular discussions of annotation disagreements and model behavior double as methodological training, expanding the pool of researchers who can lead future work on their own languages.

\paragraph{Scalable human-AI collaboration.}
In the next phase, models will propose translations for additional monolingual text, and community reviewers will accept, correct, or reject them. This human-in-the-loop paradigm scales corpus creation while maintaining quality and keeps meaningful decision-making power with the communities most affected.

\paragraph{Downstream social impact.}
Potential downstream uses under discussion with local stakeholders include multilingual interfaces for government schemes, translation support for frontline health and education workers, and digital archiving of oral narratives and traditional knowledge. Each application relies on robust MT systems and on data whose collection and use are co-governed with the communities concerned.

\section{Conclusion}
We introduce AdiBhashaa, the first open, community-curated MT benchmark
for Bhili, Mundari, Gondi, and Santali, and show that participatory corpus
creation combined with careful fine-tuning can substantially improve support
for severely under-resourced languages in contemporary models. The work
demonstrates that high-quality parallel data for such languages can be built at
moderate scale through sustained collaboration with native speakers, and that
even relatively small corpora can unlock useful performance gains in multilingual systems. We hope that AdiBhashaa will serve both as a practical resource
for MT research and as a template for similar initiatives targeting other under-
represented languages.

\bibliography{aaai2026}

@article{galaindictrans2,
  title={IndicTrans2: Towards High-Quality and Accessible Machine Translation Models for all 22 Scheduled Indian Languages},
  author={Gala, Jay and Chitale, Pranjal A and Raghavan, AK and Gumma, Varun and Doddapaneni, Sumanth and Nawale, Janki Atul and Sujatha, Anupama and Puduppully, Ratish and Raghavan, Vivek and Kumar, Pratyush and others},
  journal={Transactions on Machine Learning Research},
year={2023}
}

@article{costa2022no,
  title={No language left behind: Scaling human-centered machine translation},
  author={Costa-Juss{\`a}, Marta R and Cross, James and {\c{C}}elebi, Onur and Elbayad, Maha and Heafield, Kenneth and Heffernan, Kevin and Kalbassi, Elahe and Lam, Janice and Licht, Daniel and Maillard, Jean and others},
  journal={arXiv preprint arXiv:2207.04672},
  year={2022}
}

@inproceedings{xue2021mt5,
  title={mT5: A massively multilingual pre-trained text-to-text transformer},
  author={Xue, Linting and Constant, Noah and Roberts, Adam and Kale, Mihir and Al-Rfou, Rami and Siddhant, Aditya and Barua, Aditya and Raffel, Colin},
  booktitle={Proceedings of the 2021 conference of the North American chapter of the association for computational linguistics: Human language technologies},
  pages={483--498},
  year={2021}
}

@inproceedings{muennighoff2023crosslingual,
  title={Crosslingual generalization through multitask finetuning},
  author={Muennighoff, Niklas and Wang, Thomas and Sutawika, Lintang and Roberts, Adam and Biderman, Stella and Le Scao, Teven and Bari, M Saiful and Shen, Sheng and Yong, Zheng-Xin and Schoelkopf, Hailey and others},
  booktitle={Proceedings of the 61st Annual Meeting of the Association for Computational Linguistics (Volume 1: Long Papers)},
  pages={15991--16111},
  year={2023}
}

@article{comanici2025gemini,
  title={Gemini 2.5: Pushing the frontier with advanced reasoning, multimodality, long context, and next generation agentic capabilities},
  author={Comanici, Gheorghe and Bieber, Eric and Schaekermann, Mike and Pasupat, Ice and Sachdeva, Noveen and Dhillon, Inderjit and Blistein, Marcel and Ram, Ori and Zhang, Dan and Rosen, Evan and others},
  journal={arXiv preprint arXiv:2507.06261},
  year={2025}
}

@article{hurst2024gpt,
  title={Gpt-4o system card},
  author={Hurst, Aaron and Lerer, Adam and Goucher, Adam P and Perelman, Adam and Ramesh, Aditya and Clark, Aidan and Ostrow, AJ and Welihinda, Akila and Hayes, Alan and Radford, Alec and others},
  journal={arXiv preprint arXiv:2410.21276},
  year={2024}
}

@article{xaxa2005politics,
  title={Politics of language, religion and identity: Tribes in India},
  author={Xaxa, Virginius},
  journal={Economic and political weekly},
  pages={1363--1370},
  year={2005},
  publisher={JSTOR}
}

@misc{census2011_st15_bihar,
  author = "{Office of the Registrar General and Census Commissioner, India}",
  title  = "{Population Census 2011, Table ST-15: Scheduled Tribe by Mother Tongue (for Each Tribe Separately), Bihar}",
  year   = "2011",
  note   = "Available at https://censusindia.gov.in/nada/index.php/catalog/12542"
}

\end{document}